\pdfoutput=1

\documentclass[11pt]{article}

\usepackage[preprint]{acl}

\usepackage{times}
\usepackage{latexsym}
\usepackage{booktabs}
\usepackage{multirow} 
\usepackage{amsmath}
\usepackage{amsfonts}
\usepackage{colortbl}
\usepackage{graphicx} 
\usepackage{subcaption} 
\usepackage[T1]{fontenc}

\usepackage[utf8]{inputenc}

\usepackage{microtype}

\usepackage{inconsolata}

\usepackage{graphicx}

\title{RGAR: Recurrence Generation-augmented Retrieval for Factual-aware Medical Question Answering}

\author{
 \textbf{Sichu Liang \textsuperscript{1}}\thanks{Equal Contribution},
 \textbf{Linhai Zhang \textsuperscript{2}}\footnotemark[1],
 \textbf{Hongyu Zhu \textsuperscript{3}}\footnotemark[1],
 \textbf{Wenwen Wang\textsuperscript{4}},
 \textbf{Yulan He\textsuperscript{2, 5}},
 \textbf{Deyu Zhou \textsuperscript{1}}\thanks{Corresponding author} 
\\
 \textsuperscript{1}School of Computer Science and Engineering, Key Laboratory of New Generation Artificial Intelligence \\Technology and Its Interdisciplinary Applications, Southeast University, Ministry of Education, China
 \\
\textsuperscript{2}Department of Informatics, King's College London, UK\\
 \textsuperscript{3}	School of Electronic Information and Electrical Engineering, Shanghai Jiao Tong University, China\\
 \textsuperscript{4}School of Electrical and Computer Engineering, Carnegie Mellon University, USA\\
 \textsuperscript{5}The Alan Turing Insitute, UK\\
}

\begin{document}
\maketitle
\begin{abstract}
Medical question answering requires extensive access to specialized \textit{conceptual knowledge}. The current paradigm, Retrieval-Augmented Generation (RAG), acquires expertise medical knowledge through large-scale corpus retrieval and uses this knowledge to guide a general-purpose large language model (LLM) for generating answers. 
However, existing retrieval approaches often overlook the importance of \textit{factual knowledge}, which limits the relevance of retrieved conceptual knowledge and restricts its applicability in real-world scenarios, such as clinical decision-making based on Electronic Health Records (EHRs).
This paper introduces RGAR, a recurrence generation-augmented retrieval framework that retrieves both relevant \textit{factual} and \textit{conceptual} knowledge from dual sources (i.e., EHRs and the corpus), allowing them to interact and refine each another.
Through extensive evaluation across three factual-aware medical question answering benchmarks, RGAR establishes a new state-of-the-art performance among medical RAG systems.
Notably, the Llama-3.1-8B-Instruct model with RGAR surpasses the considerably larger, RAG-enhanced GPT-3.5. 
Our findings demonstrate the benefit of extracting factual knowledge for retrieval, which consistently yields improved generation quality.
\end{abstract}

\section{Introduction}
Large Language Models (LLMs) have demonstrated remarkable capabilities in general question answering (QA) tasks, achieving impressive performance across diverse scenarios \cite{achiam2023gpt}. However, when facing domain-specific questions that require specialized expertise, from medical diagnosis \cite{jin2021disease} to legal charge prediction \cite{wei-etal-2024-mud}, these models face significant challenges, often generating unreliable conclusions due to both hallucinations \cite{ji2023survey} and potentially stale knowledge embedded in their parameters \cite{wang2024knowledge}. 



\begin{figure}
    \centering
    \includegraphics[width=\linewidth]{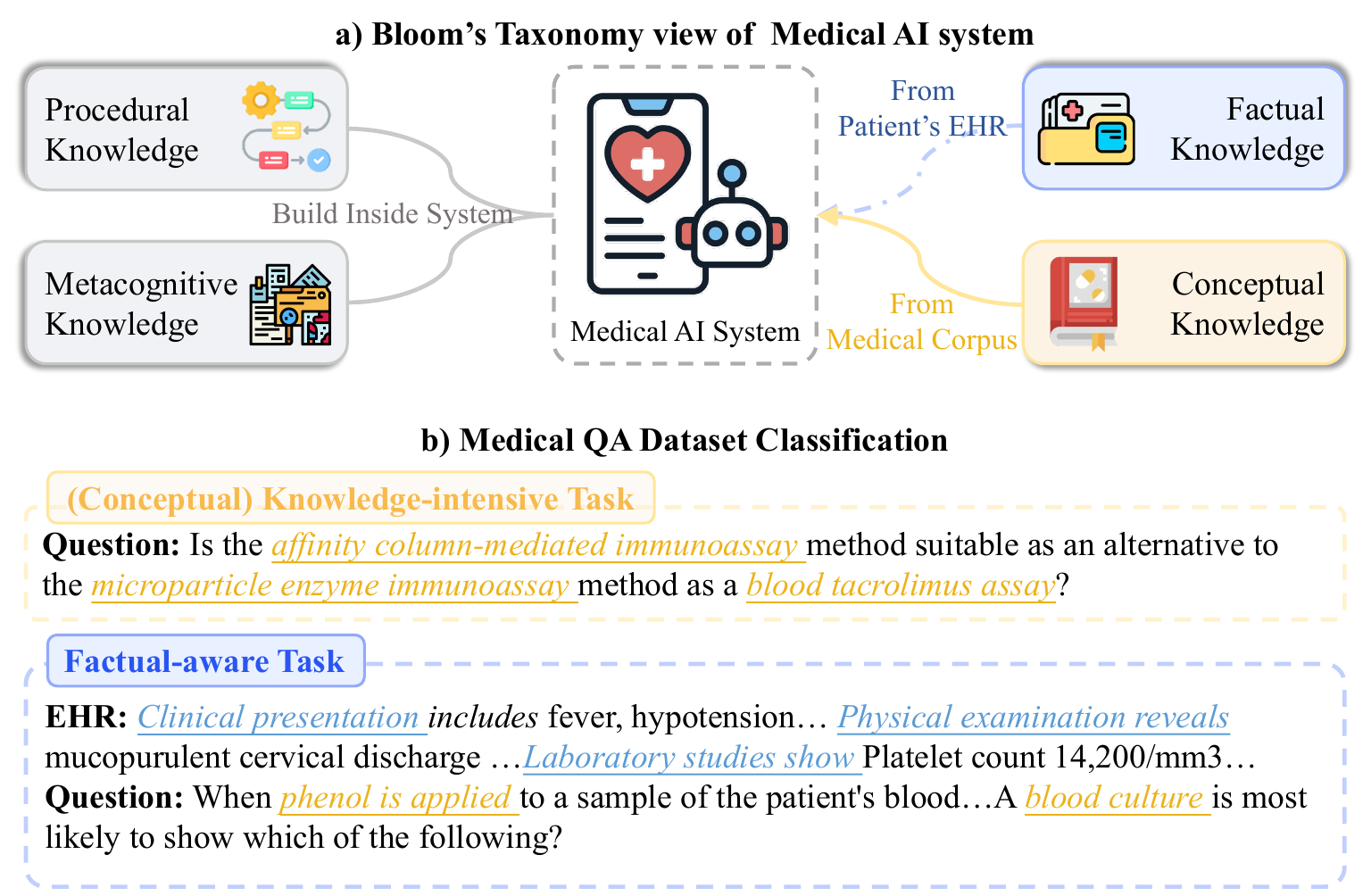}
    \caption{a) Medical AI Systems from the Perspective of Bloom's Taxonomy. b) Two Types of Medical Question Answering Tasks.}
    \label{fig:enter-label}
\end{figure}

\textbf{Retrieval-Augmented Generation (RAG)} \cite{lewis2020retrieval} has emerged as a promising approach to address these challenges by leveraging extensive, trustworthy knowledge bases to support LLM reasoning. The effectiveness of this approach, however, heavily depends on the relevance of retrieved documents. Recent advances, such as \textbf{Generation-Augmented Retrieval (GAR)} \cite{mao-etal-2021-generation}, focus on enhancing retrieval performance by generating relevant context for query expansion.


In the medical domain, current RAG approaches concatenate all available contextual information from a given example into a single basic query for retrieval, aiming to provide comprehensive context for model reasoning \cite{xiong-etal-2024-benchmarking}. While this method has demonstrated substantial improvements on early \textit{knowledge-intensive} medical QA datasets such as PubMedQA \cite{jin-etal-2019-pubmedqa}, its limitations have become increasingly apparent with the emergence of EHR-integrated datasets that better reflect real-world clinical practices \cite{kweon2024ehrnoteqa}. Electronic Health Records (EHRs) typically contain extensive patient data, including comprehensive diagnostic test results and medical histories \cite{pang2021cehr}. However, for any specific medical query, only a small subset of this information is typically relevant, and retrieval performance can be significantly degraded when queries are diluted with extraneous EHR content \cite{johnson2023mimic, lovon-melgarejo-etal-2024-revisiting}.



We highlight that current \textit{retrieval methods} often fail to adequately consider \textit{factual information} in real-world medical scenarios. Crucially, even when applying query expansion with GAR, the persistent oversight of factual information fundamentally limits their ability to retrieve real relevant documents.



Inspired by \textbf{Bloom's taxonomy} \cite{forehand2010bloom,markus2001toward}, we categorize the knowledge required to address real-world medical QA problems into four types: \textit{Factual Knowledge}, \textit{Conceptual Knowledge}, \textit{Procedural Knowledge}, and \textit{Metacognitive Knowledge}.
The latter two represent higher-order knowledge typically embedded within advanced RAG systems. Specifically, \textit{Procedural Knowledge} refers to the processes and strategies required to solve problems, such as problem decomposition and retrieval \cite{wei2022chain, zhou2023leasttomost}, while \textit{Metacognitive Knowledge} pertains to an LLM's ability to assess whether it has sufficient knowledge or evidence to perform effective reasoning \cite{kim-etal-2023-tree, wang-etal-2023-self-knowledge}.

\textit{Factual Knowledge} and \textit{Conceptual Knowledge} require retrieval from large databases containing substantial amounts of irrelevant content, corresponding to the EHRs of patients and medical corpora in answering medical questions. Unfortunately, current RAG systems do not differentiate between these types of \textit{retrieval targets}, overlooking the necessity of retrieval from EHRs.


To overcome this limitation, we propose \textbf{RGAR}, a system designed to simultaneously retrieves \textit{Factual Knowledge} and \textit{Conceptual Knowledge} through a recurrent query generation and interaction mechanism. This approach iteratively refines queries to enhance the relevance of retrieved professional and factual knowledge, thereby improving performance on \textit{knowledge-intensive} and \textit{factual-aware} medical QA tasks.

Our key contributions are listed as follows:
\begin{itemize}
\item We are the first to analyze RAG systems through the lens of Bloom's taxonomy, addressing the current underrepresentation of \textit{Factual Knowledge} in existing frameworks.
\item We introduce RGAR, a dual-end retrieval system that facilitates recurrent interactions between \textit{Factual} and \textit{Conceptual} Knowledge, bridging the gap between LLMs and real-world clinical applications.
\item Through extensive experiments on three medical QA datasets involving \textit{Factual Knowledge}, we demonstrate that RGAR achieves superior average performance compared to state-of-the-art (SOTA) methods, enabling Llama-3.1-8B-Instruct model to outperform the considerably larger RAG-enhanced GPT-3.5-turbo.
\end{itemize}

\section{Related Work}
\textbf{RAG Systems. } RAG systems are characterized as a "Retrieve-then-Read" framework \cite{gao2023retrieval}. The development of Naive RAG has primarily focused on retriever optimization, evolving from discrete retrievers such as BM25 \cite{friedman1977algorithm} to more sophisticated and domain-specific dense retrievers, including DPR \cite{karpukhin-etal-2020-dense} and MedCPT \cite{jin2023medcpt}, which demonstrate superior performance.

In recent years, numerous advanced RAG systems have emerged. Advanced RAG systems focus on designing multi-round retrieval structures, including iterative retrieval \cite{sun2019pullnet}, recursive retrieval \cite{sarthi2024raptor}, and adaptive retrieval \cite{jeong-etal-2024-adaptive}. A notable work in medical QA is MedRAG \cite{xiong-etal-2024-benchmarking}, which analyzes retrievers, corpora, and LLMs, offering practical guidelines. Follow-up work, $i$-MedRAG \cite{xiong2024improving}, improved performance through multi-round decomposition and iteration, albeit with significant computational costs.

These approaches focus solely on optimizing the retrieval process, overlooking the retrievability of \textit{factual knowledge}. In contrast, RGAR introduces a recurrent structure, enabling continuous query optimization through dual-end retrieval and extraction from EHRs and professional knowledge corpora, thereby enhancing access to both knowledge types.

\textbf{Query Optimization. } As the core interface in human-AI interaction, query optimization (also known as prompt optimization) is the key to improving AI system performance. It is widely applied in tasks such as text-to-image generation \cite{liu2022compositional, wu-etal-2024-universal} and code generation \cite{nazzal2024promsec}.

In the era of large language models, query optimization for retrieval tasks has gained increasing attention. Representative work includes GAR \cite{mao-etal-2021-generation}, which improves retrieval performance through query expansion using fine-tuned BERT models \cite{devlin-etal-2019-bert}. GENREAD \cite{yu2023generate} further explored whether LLM-generated contexts could replace retrieved professional documents as reasoning evidence. MedGENIE \cite{frisoni-etal-2024-generate} extended this approach to medical QA.

Another line of work focuses on query transformation and decomposition, breaking down original queries into multiple sub-queries tailored to specific tasks, enhancing retrieval alignment with model needs \cite{dhuliawala2023chain}. Subsequent work has reinforced the effectiveness of query decomposition through fine-tuning \cite{ma2023query}.

Using expanded queries directly as reasoning evidence lacks the transparency of RAG, as RAG relies on retrievable documents that provide traceable and trustworthy reasoning, which is crucial in the medical field.
Besides, the effectiveness of query expansion and query decomposition approaches is heavily dependent on fine-tuning LLMs, which limits scalability.


In contrast, our work focuses on query optimization without fine-tuning LLMs. Specifically, retrieval from EHRs can be seen as query filtering that eliminates irrelevant information, thereby obtaining pertinent \textit{factual knowledge}. Extracting factual knowledge enhances the effectiveness of retrieval from the corpus.





\begin{figure*}
    \centering
    \includegraphics[width=\linewidth]{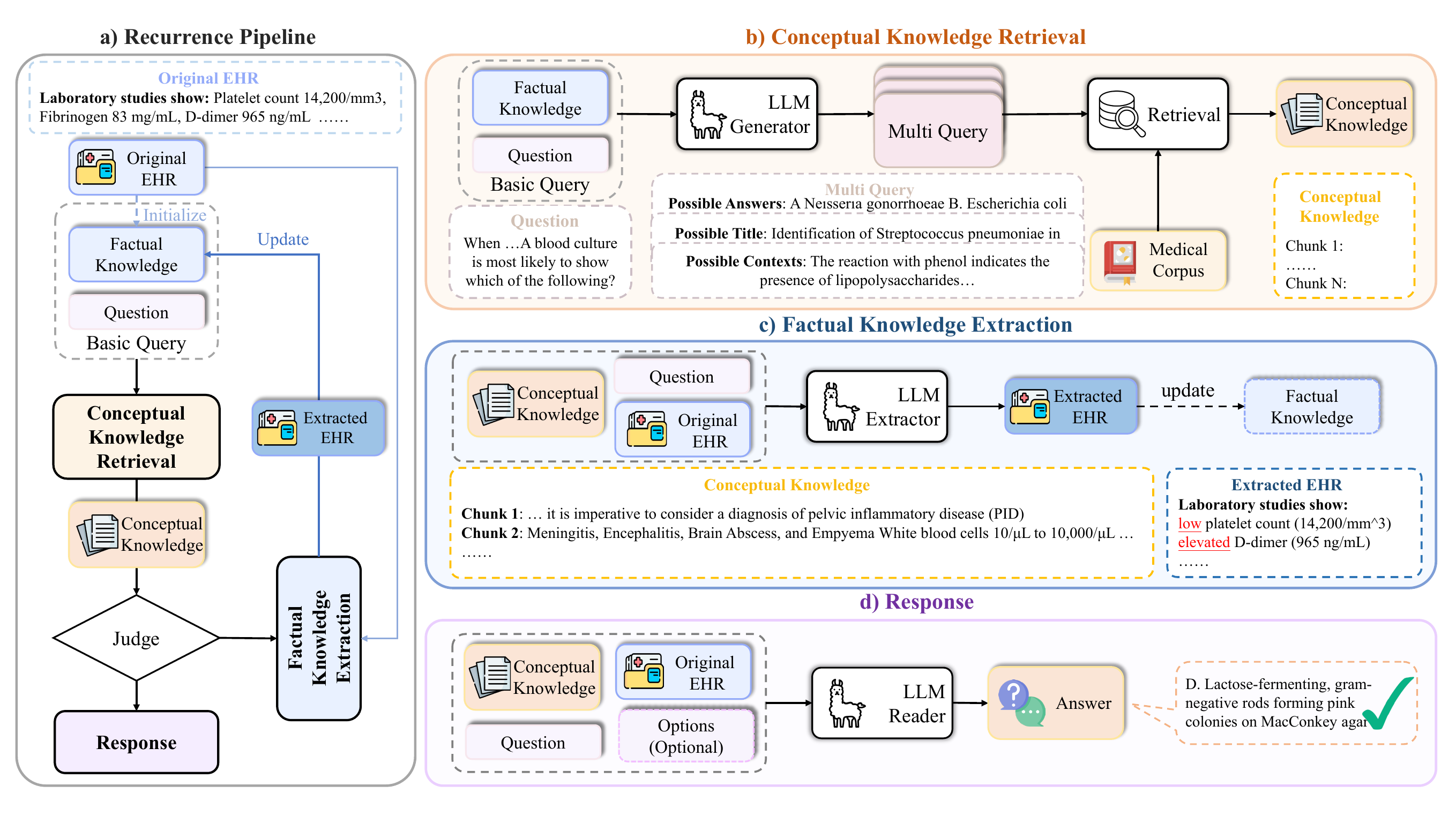}
    \caption{The Overall Framework of RGAR. a) The Recurrence Pipeline in § \ref{sec:pipeline}; b) Conceptual Knowledge Retrieval in § \ref{sec:Train-free}; c) Factual Knowledge Extraction in § \ref{sec:Extraction}; d) Response Template in § \ref{sec:pipeline}.}
    \label{fig:pipeline}
\end{figure*}

\section{Methodology}
In this section, we introduce RGAR framework, as illustrated in Figure \ref{fig:pipeline}. It begins by prompting a general-purpose LLM to generate multiple queries from an initial basic query. These multiple queries are then used to \textbf{retrieve conceptual knowledge} from the corpus (§ \ref{sec:Train-free}). Then retrieved conceptual knowledge is subsequently used to \textbf{extract factual knowledge} from the electronic health records (EHRs) and transform it into retrieval-optimized representations (§ \ref{sec:Extraction}). The \textbf{recurrence pipeline} continuously updates the basic query and iteratively executes the two aforementioned components. This process optimizes the retrieved results, ultimately improving the quality of responses.(§ \ref{sec:pipeline}).
\subsection{Task Formulation}
In \textit{factual-aware} medical QA, each data sample comprises the following elements: a patient's natural language query $\mathcal{Q}$, the electronic health record (EHR) as factual knowledge $\mathcal{F}$, and a set of candidate answer $\mathcal{A} = \{a_1, ..., a_{|\mathcal{A}|}\}$. The overall goal is to identify the correct answer $\hat{a}$ from $\mathcal{A}$.

A \textit{non-retrieval} approach directly prompts an LLM to act as a \textbf{reader}, processing the entire context and generating an answer, formulated as:

\begin{equation}
\hat{a}=\textbf{LLM}(\mathcal{F},\mathcal{Q},\mathcal{A}|\mathcal{T}_r)
\end{equation}

where $\mathcal{T}_r$ is the prompts. However, this approach relies exclusively on the conceptual knowledge encoded within LLM, without leveraging external, trustworthy medical knowledge sources.

To overcome this limitation, recent studies have explored \textit{retrieval-based} approaches, which enhance the model’s knowledge by retrieving a specified number $N$ of chunks, denoted as $\mathcal{C} = \{c_1, ..., c_N\}$, from a chunked corpus (knowledge base) $\mathcal{K}$. This answering process is expressed as:

\begin{equation}
\hat{a}=\textbf{LLM}(\mathcal{F},\mathcal{Q},\mathcal{A},\mathcal{C}|\mathcal{T}_r).
\label{eq:retrieval-augmented}
\end{equation}

\subsection{Conceptual Knowledge Retrieval (CKR)}
\label{sec:Train-free}
To maintain consistency with the \textit{option-free retrieval approach} proposed by \cite{xiong-etal-2024-benchmarking}, we do not incorporate the answer options $\mathcal{A}$ during retrieval. This design is in line with real-world medical quality assurance scenarios, where answer choices are typically not available in advance.

Following their method, we construct the \textbf{basic query} by concatenating the EHR and the patient's query, formally defined as $q_b = \mathcal{Q} \oplus \mathcal{F}$, where $\oplus$ denotes text concatenation.

Traditional dense retrievers, such as Dense Passage Retrieval (DPR) \cite{karpukhin-etal-2020-dense}, identify the top-$N$ relevant chunks $C$ from the knowledge base $\mathcal{K}$ by computing similarity scores using an encoder $E$:

\begin{equation}
\begin{split}
    &\text{sim}(q_b, c_i) = E(q_b)^\top E(c_i), \\
    &\mathcal{C} = \text{top-}N(\{\text{sim}(q_b, c_i)\}).
\end{split}
\end{equation}

Vanilla GAR \cite{mao-etal-2021-generation} expands $q_b$ using a fine-tuned BERT \cite{devlin-etal-2019-bert} to produce three types of content that enhance retrieval: potential answers $q_e^a$, contexts $q_e^c$, and titles $q_e^t$.
With the growing zero-shot generation capabilities of LLMs \cite{kojima2022large}, a common practice is to prompt LLMs to serve as train-free query \textbf{generators}, producing expanded content $\tilde{q}_e$ using prompt templates $\mathcal{T}_g$ \cite{frisoni-etal-2024-generate}. The three types of content generation process can be formulated as:

\begin{equation}
\label{eq:query-generation}
\begin{array}{l}
\tilde{q}_e^a = \textbf{LLM}(q_b |\mathcal{T}^a_g), \\[1ex]
\tilde{q}_e^c = \textbf{LLM}(q_b |\mathcal{T}^c_g), \\[1ex]
\tilde{q}_e^t = \textbf{LLM}(q_b |\mathcal{T}^t_g).
\end{array}
\end{equation}

The final score $Sc$ for retrieving $\mathcal{C}$ is then computed by normalizing and averaging the similarities of these expanded queries:
\begin{equation}
\label{eq:normalized-retrieval-score}
\text{Sc}(c_i) = \sum_{\tilde{q}_e \in \{\tilde{q}_e^a, \tilde{q}_e^c, \tilde{q}_e^t\}} \frac{\exp(\text{sim}(\tilde{q}_e, c_i))}{\sum_{c_j} \exp(\text{sim}(\tilde{q}_e, c_j))}.
\end{equation}

\subsection{Factual Knowledge Extraction (FKE)}
\label{sec:Extraction}

In EHR, only a small portion of necessary information constitutes problem-relevant factual knowledge \cite{d2004evaluation}. Direct input of lengthy EHR content containing substantial irrelevant information into dense retrievers can degrade retrieval performance \cite{ren-etal-2023-thorough}. While a straightforward approach would be to retrieve EHR content based on question $\mathcal{Q}$ \cite{factual_aware}, this fails to fully utilize conceptual knowledge obtained from previous Conceptual Knowledge Retrieval Stage. Furthermore, the necessary chunking of EHR for retrieval introduces content discontinuity \cite{luo-etal-2024-landmark}.

Given that EHRs more closely resemble long passages from the Needle in a Haystack task \cite{kamradt2024needle} rather than necessarily chunked corpus, and inspired by large language models' capability to precisely locate answer spans in reading comprehension tasks \cite{cheng2024adapting}, we propose leveraging LLMs for text span tasks \cite{rajpurkar-etal-2016-squad} on EHR to filter relevant factual knowledge efficiently and effectively using conceptual knowledge. We define this filtered factual knowledge as $\mathcal{F}_s$, with prompts $\mathcal{T}_s$, expressed as:
\begin{equation}
    \mathcal{F}_s=\textbf{LLM}(\mathcal{F},\mathcal{Q},\mathcal{C}|\mathcal{T}_s).  
\end{equation}

In addition, EHRs often contain numerical report results \cite{lovon-melgarejo-etal-2024-revisiting} that require conceptual knowledge to interpret their significance. Furthermore, medical QA involves multi-hop questions \cite{pal2022medmcqa}, where retrieved conceptual knowledge can generate explainable new factual knowledge conducive to reasoning. Drawing from LLM zero-shot summarization prompting strategies \cite{wu2025towards}, we analyze and summarize the filtered EHR $\mathcal{F}_s$ with prompts $\mathcal{T}_e$, yielding an enriched representation $\mathcal{F}_e$:
\begin{equation}
    \mathcal{F}_e=\textbf{LLM}(\mathcal{F}_s,\mathcal{Q},\mathcal{C}|\mathcal{T}_e).  
\end{equation}

This process, which we refer to as the LLM \textbf{Extractor}, completes the extraction of original EHR information. In practice, RGAR implements these two phases using single-stage prompting to reduce time overhead.

\subsection{The Recurrence Pipeline and Response}
\label{sec:pipeline}

Building on the \(\mathcal{F}_e\), we \textbf{update} the basic query for Conceptual Knowledge Retrieval as \(q_b = \mathcal{Q} \oplus \mathcal{F}_e\). This establishes a \textbf{recurrence interaction} between factual and conceptual knowledge, guiding next retrieval toward more relevant content. Iterative execution enhances the stability of both retrieval and extraction. The entire pipeline recurs for a predefined number of iterations, ultimately yielding the final retrieved conceptual knowledge $\mathcal{C}^*$.

During the response phase, we follow the approach in Equation \ref{eq:retrieval-augmented} to generate answers. Notably, the $\mathcal{F}_e$ are restricted to the retrieval phase and are not used in the response phase. The sole difference lies in the retrieved chunks, highlighting the impact of retrieval quality on the responses.

\section{Experiments}
\subsection{Experimental Setup}
\subsubsection{Benchmark Datasets}

We evaluated RGAR on three \textit{factual-aware} medical QA benchmarks featuring multiple-choice questions that require human-level reading comprehension and expert reasoning to analyze patients' clinical conditions.



\textbf{MedQA-USMLE} \cite{jin2021disease} and \textbf{MedMCQA} \cite{pal2022medmcqa} consist of questions derived from professional medical exams, evaluating specialized expertise such as disease symptom diagnosis and medication dosage requirements. The problems frequently involve patient histories, vital signs (e.g., blood pressure, temperature), and final diagnostic evaluations (e.g., CT scans), making it necessary to retrieve relevant medical knowledge tailored to the patient’s specific circumstances. However, due to their exam-oriented format, the provided information has already been filtered, reducing the difficulty of extracting factual knowledge from EHR.

\textbf{EHRNoteQA} \cite{kweon2024ehrnoteqa} is a recently introduced benchmark that provides authentic, complex EHR data derived from MIMIC-IV \cite{johnson2023mimic}. This dataset encompasses a wide range of topics and demands that models emulate genuine clinical consultations, ultimately generating accurate discharge recommendations. Consequently, EHRNoteQA challenges models to identify which \textit{factual details} within the EHR are relevant to the questions at hand and apply domain-specific knowledge to address them.

\begin{table}[htbp]
  \centering
  \caption{Medical QA Benchmark Statistics.}
  \resizebox{\linewidth}{!}{ 
    \begin{tabular}{lccc}
      \toprule
      Benchmarks & Max. Len & Avg. Len & Min. Len \\
            \midrule
      \rowcolor{gray!20} \multicolumn{4}{c}{Non-EHR QA Benchmarks} \\
      \midrule
        BioASQ-Y/N & 52 & 17 & 9  \\
      PubMedQA & 57 & 23 & 10  \\
      \midrule
      \rowcolor{gray!20} \multicolumn{4}{c}{EHR QA Benchmarks} \\
      \midrule
      MedMCQA & 207 & 41 & 11  \\
      MedQA-USMLE & 872 & 197 & 50  \\
      EHRNoteQA & 5782 & 3061 & 667  \\

      \bottomrule
    \end{tabular}
  }
  \label{tab:qa_benchmarks}
\end{table}

Table \ref{tab:qa_benchmarks} highlights that the chosen datasets, which include EHR information, tend to have significantly \textbf{longer} content compared to datasets without EHRs. Notably, the EHRNoteQA dataset has a maximum length exceeding 4,000 tokens. This raises concerns about the reasonableness of directly employing these EHRs for retrieval.

\begin{table*}[htbp]
  \centering
  \caption{Comparison of RGAR with Other Methods on Three Factual-Aware Datasets. $\Delta$ Indicates Improvement Over Custom, \textbf{Bold} Represents the Best, and \underline{Underline} Indicates the Second-Best.}
  \resizebox{\linewidth}{!}{%
    \begin{tabular}{llcccccc|cc}
      \toprule
      \multicolumn{2}{c}{\multirow{2}{*}{Method}} & \multicolumn{2}{c}{MedQA-USMLE (\# 1273)} & \multicolumn{2}{c}{MedMCQA(\# 4183)} & \multicolumn{2}{c|}{EHRNoteQA(\# 962)} & \multicolumn{2}{c}{Average(↓)} \\
      \cmidrule(lr){3-4} \cmidrule(lr){5-6} \cmidrule(lr){7-8} \cmidrule(lr){9-10}
      \multicolumn{2}{c}{} & Acc. & $\Delta$ & Acc. & $\Delta$ & Acc. & $\Delta$ & Acc. & $\Delta$ \\
      \midrule
      \multirow{2}{*}{w/o Retrieval} & Custom  & 50.20 & 0.00  & 50.01 & 0.00  & 47.19 & 0.00  & 49.13 & 0.00  \\
                               & CoT     & 51.45 & 1.25  & 44.53 & -5.48 & 62.89 & 15.70 & 52.96 & 3.82  \\
      \midrule
      \multirow{5}{*}{w/ Retrieval}  & RAG     & 53.50 & 3.30  & \underline{50.54} & \underline{0.53}  & 61.12 & 13.93 & 55.05 & 5.92  \\
                               & MedRAG  & 50.27 & 0.07  & 47.53 & -2.48 & 70.58 & 23.39 & 56.13 & 6.99  \\
                               & GAR     & \underline{57.97} & \underline{7.77}  & 50.42 & 0.41  & 65.48 & 18.29 & 57.96 & 8.82  \\
                               & $i$-MedRAG & 56.24 & 6.04  & 44.94 & -5.07 & \textbf{74.22} & \textbf{27.03} & \underline{58.47} & \underline{9.33}  \\
                               & RGAR    & \textbf{58.83} & \textbf{8.63}  & \textbf{51.02} & \textbf{1.01}  & \underline{73.28} & \underline{26.09} & \textbf{61.04} & \textbf{11.91} \\
      \bottomrule
    \end{tabular}%
  }
  \label{tab:mian_results}
\end{table*}


\subsubsection{Retriever and Corpus}
To ensure a fair comparison, we adopt the same retriever, corpus, and parameter settings as previous work \cite{xiong-etal-2024-benchmarking}. We use MedCPT \cite{jin2023medcpt}, a dense retriever specialized for the biomedical domain, configured to retrieve 32 chunks by default. For the corpus, we employ the Textbooks dataset \cite{jin-etal-2019-pubmedqa}, a lightweight collection of 125.8k chunks derived from medical textbooks, with an average length of 182 tokens.

\subsubsection{LLMs and Baselines}
We focus on the effect of RGAR on general-purpose LLMs without domain-specific knowledge. Therefore, we exclude LLMs fine-tuned on the medical domain, such as PMC-Llama \cite{wu2024pmc}. 
Our primary experiments utilize Llama-3.2-3B-Instruct, while ablation studies include a range of models from the Llama-3.1/3.2 \cite{dubey2024llama} and Qwen-2.5 \cite{yang2024qwen2} families, ranging from 1.5B to 8B parameters. All selected models feature a context length of approximately 128K tokens.
Temperatures are set to zero to ensure reproducibility through greedy decoding. 

For \textit{non-retrieval methods}, we consider a zero-shot approach Custom \cite{kojima2022large} as a baseline and evaluate improvements relative to it. To fully exploit the reasoning capabilities of the LLMs, we incorporate chain-of-thought (CoT) reasoning \cite{wei2022chain}.
For \textit{retrieval-based methods}, we evaluate the classic RAG model \cite{lewis2020retrieval}, the domain-adapted MedRAG \cite{xiong-etal-2024-benchmarking}, and $i$-MedRAG \cite{xiong2024improving}, a medical-domain RAG system designed to decompose questions and iteratively provide answers.

We adopt GAR \cite{mao-etal-2021-generation} as a representative \textit{query-optimized RAG method}, implemented train-free in accordance with § \ref{sec:Train-free}. RGAR defaults to \textbf{2} rounds of recurrence.

\subsubsection{Evaluation Settings}
Following MIRAGE \cite{xiong-etal-2024-benchmarking}, we adopt the following evaluation framework. In \textbf{Option-Free Retrieval}, no answer options are provided for retrieval (§\ref{sec:Train-free}), ensuring a more realistic medical QA scenario. In \textbf{Zero-Shot Learning}, RAG systems are evaluated without in-context few-shot learning, reflecting the lack of similar exemplars in real-world medical questions. For \textbf{Metrics}, we employ Accuracy, defined as the proportion of correctly answered questions, and we extract model outputs by applying regular expression matching to the entire generated responses \cite{wang-etal-2024-answer-c}.


\subsection{Main Results}
\subsubsection{Cross-Dataset Performance Improvement}
\label{cross-dataset}
\begin{figure*}[ht]
    \centering
    \begin{subfigure}{0.32\textwidth}
        \includegraphics[width=\textwidth]{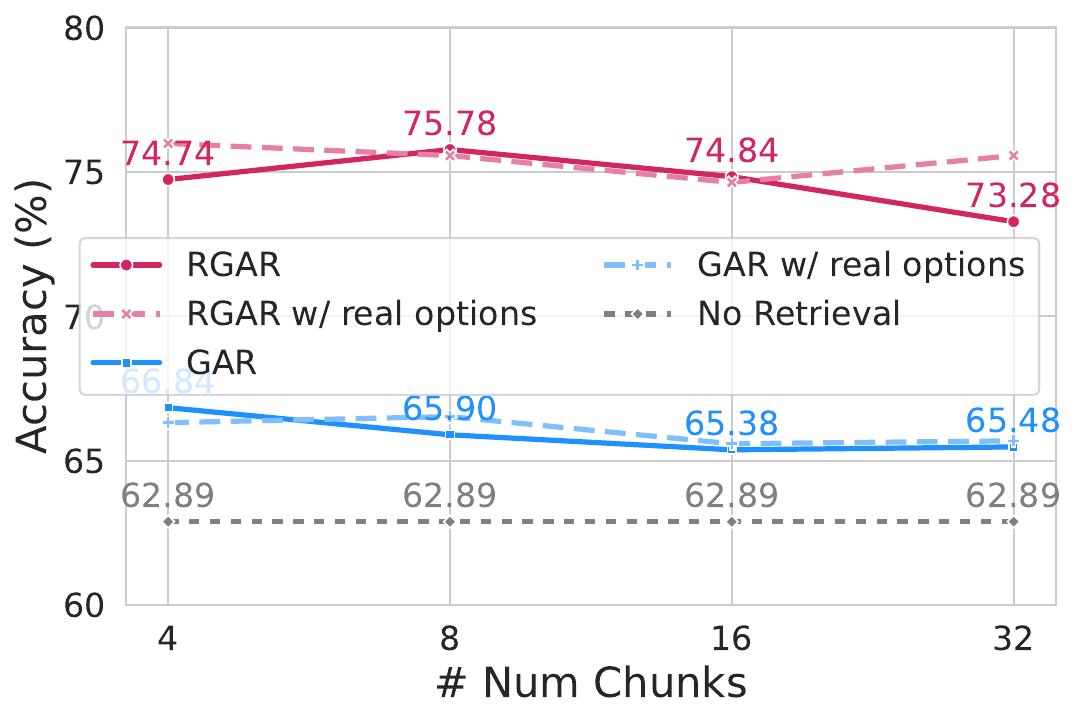}
        \caption{Effect of Using Original Options.}
        \label{fig:sub1}
    \end{subfigure}
    \hfill
    \begin{subfigure}{0.32\textwidth}
        \includegraphics[width=\textwidth]{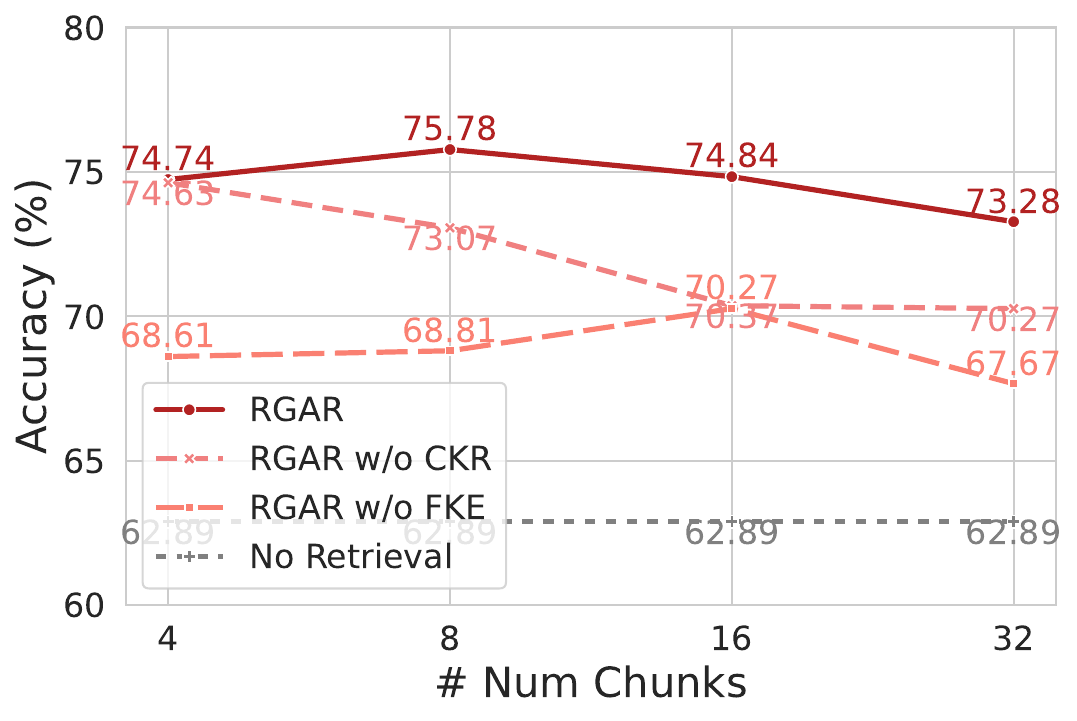}
        \caption{Effect of RGAR's Two Components.}
        \label{fig:sub2}
    \end{subfigure}
    \hfill
    \begin{subfigure}{0.32\textwidth}
        \includegraphics[width=\textwidth]{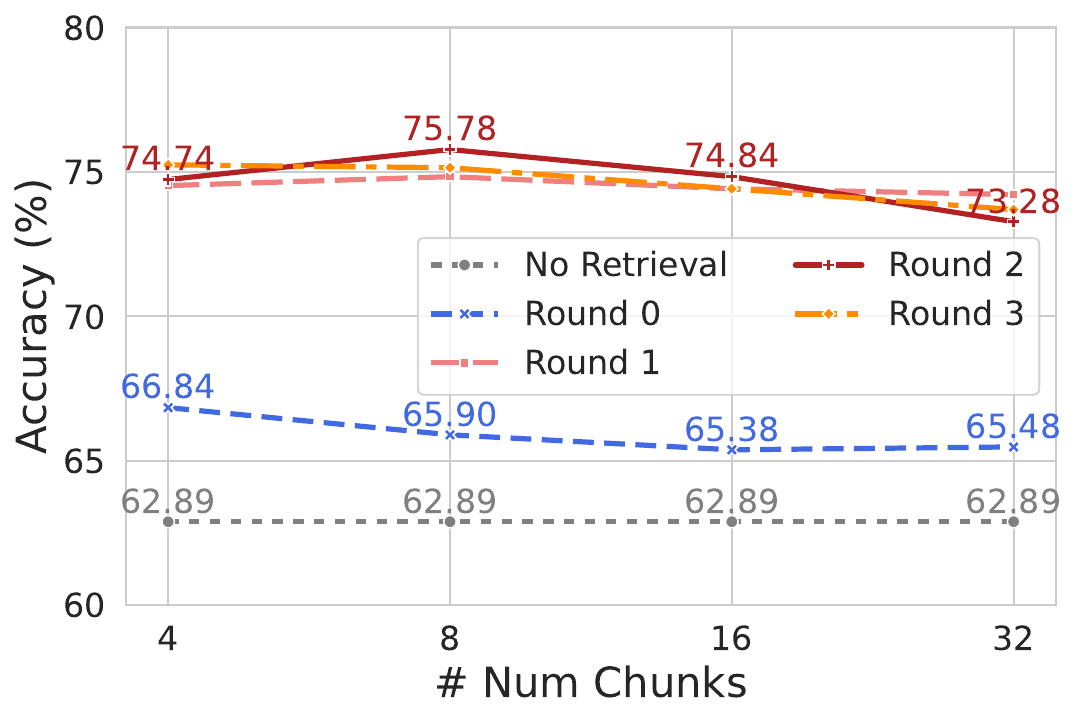}
        \caption{Effect of Rounds in RGAR.}
        \label{fig:sub3}
    \end{subfigure}
    \caption{Accuracy with Different Numbers of Retrieved Chunks on EHRNoteQA Dataset.}
    \label{fig:three_sub_figures}
\end{figure*}

We evaluate RGAR with the Llama-3.2-3B-Instruct across three factual-aware medical datasets, comparing it with several competitive baselines. Table~\ref{tab:mian_results} presents the results of all methods, along with their relative improvements over the Custom baseline. RGAR achieves the highest average performance across the three datasets, surpassing the second-best method, $i$-MedRAG, by 2\%. The retrieval-based methods, even the lowest-performing RAG, consistently outperform the non-retrieval methods Custom and CoT. This highlights the importance of retrieving specialized medical knowledge when using general-purpose LLMs to answer professional medical queries. Comparing different retrieval methods, GAR outperforms vanilla RAG by approximately 3\% on average, with a maximum improvement of 4.37\% across datasets. This indicates that generating multiple queries for retrieval provides consistent benefits. However, while performing well on EHRNoteQA, MedRAG demonstrates a negative effect on the other two datasets compared to vanilla RAG.

Notably, the improvements achieved by our RGAR over GAR exhibit a positive correlation with the average length of the dataset’s context. On EHRNoteQA, which has an average context length exceeding 3000 tokens, our approach achieved a 7.8\% improvement. This validates the advantage of our \textit{Factual knowledge Extraction} in enhancing retrieval effectiveness. Consequently, our method is particularly well-suited to real-world scenarios where complete electronic health records must be analyzed to provide medical advice. This indicates that our approach is promising for real-life applications in assisting physicians with clinical recommendations.

When analyzing performance across different datasets, we find that retrieval-based methods perform significantly better on MedQA-USMLE and EHRNoteQA, while MedMCQA showa a negative effect—consistent with results reported by MedRAG \cite{xiong-etal-2024-benchmarking}. A closer analysis reveals that MedMCQA incorporates arithmetic reasoning questions (roughly 7\% of the total), and the addition of extensive retrieved contexts diminishes the model’s numerical reasoning capabilities, which could potentially be fixed with larger base LLMs \cite{mirzadeh2025gsmsymbolic}. Nonetheless, among retrieval-based methods, our RGAR stands out as the only approach that outperforms vanilla RAG on this dataset, delivering an improvement of more than 1\% over Custom.
On EHRNoteQA, while RGAR’s performance is slightly below that of $i$-MedRAG, \textbf{the latter’s inference time is approximately 4 times longer, establishing RGAR as a more efficient and cost-effective alternative}.

\subsubsection{Base LLMs with Different Sizes and Model Families}
\begin{table}[htbp]
  \centering
  \caption{Comparison of LLMs on MedQA-USMLE.}
  \resizebox{\linewidth}{!}{ 
    \begin{tabular}{lcccc}
      \toprule
      Model & \multicolumn{1}{c}{Custom} & \multicolumn{1}{c}{RAG} & \multicolumn{1}{c}{GAR} & \multicolumn{1}{c}{RGAR} \\
      \midrule

      Llama-3.2-1B-Instruct & 38.96 & 29.30 & 30.79 & 29.85 \\
      Llama-3.2-3B-Instruct & 50.20 & 53.50 & 57.97 & 58.83 \\
      Llama-3.1-8B-Instruct & 60.80 & 62.14 & 67.39 & 69.52 \\
        \midrule
      Qwen2.5-1.5B-Instruct & 43.99 & 41.48 & 43.42 & 42.58 \\
      Qwen2.5-3B-Instruct & 48.23 & 49.96 & 53.50 & 54.28 \\
      Qwen2.5-7B-Instruct & 59.46 & 58.83 & 63.39 & 63.86 \\
      \midrule
      Average   & 50.27 & 49.20 & 52.74 & 53.15 \\
      \bottomrule
    \end{tabular}
  }
  \label{tab:performance}
\end{table}
To further assess the versatility of RGAR, we conduct evaluations on MedQA-USMLE, a widely used medical dataset, by utilizing base LLMs of various sizes and model families, specifically from Llama and Qwen. The results in Table \ref{tab:performance} show that RGAR consistently achieves the best average performance.

When considering model size, we find that retrieval-based approaches fall short of the non-retrieval Custom baseline for smaller models, such as Llama-3.2-1B-Instruct and Qwen2.5-1.5B-Instruct. These smaller models, constrained by their weaker performance, are not well-suited to leverage retrieval-enhanced information. As the model size increases, however, all retrieval-enhanced approaches exhibit notable performance gains, with RGAR yielding the most significant improvements. This trend becomes particularly pronounced for larger models. For example, RGAR achieves a 7.38\% improvement over RAG on Llama-8B, 5.33\% on Llama-3B, 5.03\% on Qwen-8B, and 4.32\% on Qwen-3B.

Moreover, we find that under the same experimental conditions, \textbf{Llama-3.1-8B-Instruct achieves a performance of 69.52\% with RGAR, surpassing the 66.22\% reported by MedRAG for GPT-3.5-16k-0613} \cite{achiam2023gpt}. This significant improvement underscores the practicality of using well-optimized retrieval methods with smaller models, enabling performance rivals those of proprietary large-scale foundational models in real-world medical recommendation tasks.

\subsection{Ablation Study}

Due to the absence of ground-truth retrieval chunks, we evaluate retrieval effectiveness through QA performance, systematically varying the number of retrieved chunks \(N\) from 4 to 32. A reduced retrieval number serves as a more stringent assessment of retrieval quality. We investigate three primary factors in Figure \ref{fig:three_sub_figures}: the effect of options generated by GAR versus those originally provided by the dataset, the contributions of CKR and FKE components, and the impact of RGAR’s recurrence rounds.

We first compare the retrieval performance between LLM-generated options and original dataset options. Figure \ref{fig:sub1} shows how RGAR and GAR perform across different values of \(N\). Both approaches maintain stable performance across different \(N\), indicating reliable retrieval quality. While using original options shows slightly higher average Accuracy, the difference is minimal. This suggests that even when GAR generates options that differ from the originals, it achieves similar retrieval results as long as the core topics align. 

We then examine the impact of RGAR's two main components—CKR and FKE—as shown in Figure \ref{fig:sub2}. When we remove the conceptual knowledge interaction from the FKE phase, the system shows only moderate improvements when extracting factual knowledge from EHR without conceptual knowledge, demonstrating the importance of integrating both types of knowledge. 
Removing the multi-query generation step from CKR causes performance to degrade as \(N\) increases, indicating that multiple queries are necessary to maintain stable retrieval.

Finally, we analyze the effect of rounds in RGAR (Round 0 means GAR), as illustrated in Figure \ref{fig:sub3}. Our results show that even a single iteration significantly improves performance by enabling interaction between factual and conceptual knowledge. Multiple rounds work similarly to a reranking mechanism \cite{mao-etal-2021-reader}, improving the ranking of important chunks and showing substantial gains even with relatively small \(N\). With \(N = 8\) , the default two-round setup achieves a performance of 75.78\%, almost 1\% better than using a single round. However, adding more rounds shows no clear benefits, as they tend to generate multi-hop factual knowledge during the FKE phase, leading CKR to retrieve multi-hop conceptual knowledge, which may cause LLMs to over-infer ~\cite{yang-etal-2024-large-language-models}. Given that each round involves one reasoning step from both the LLM extractor and LLM query generator, two rounds sufficiently support multi-hop reasoning needs \cite{lv-etal-2021-multi}.

\subsection{Fine-Grained Performance Analysis}

\begin{figure}[htbp]
    \centering
    \includegraphics[width=1\linewidth]{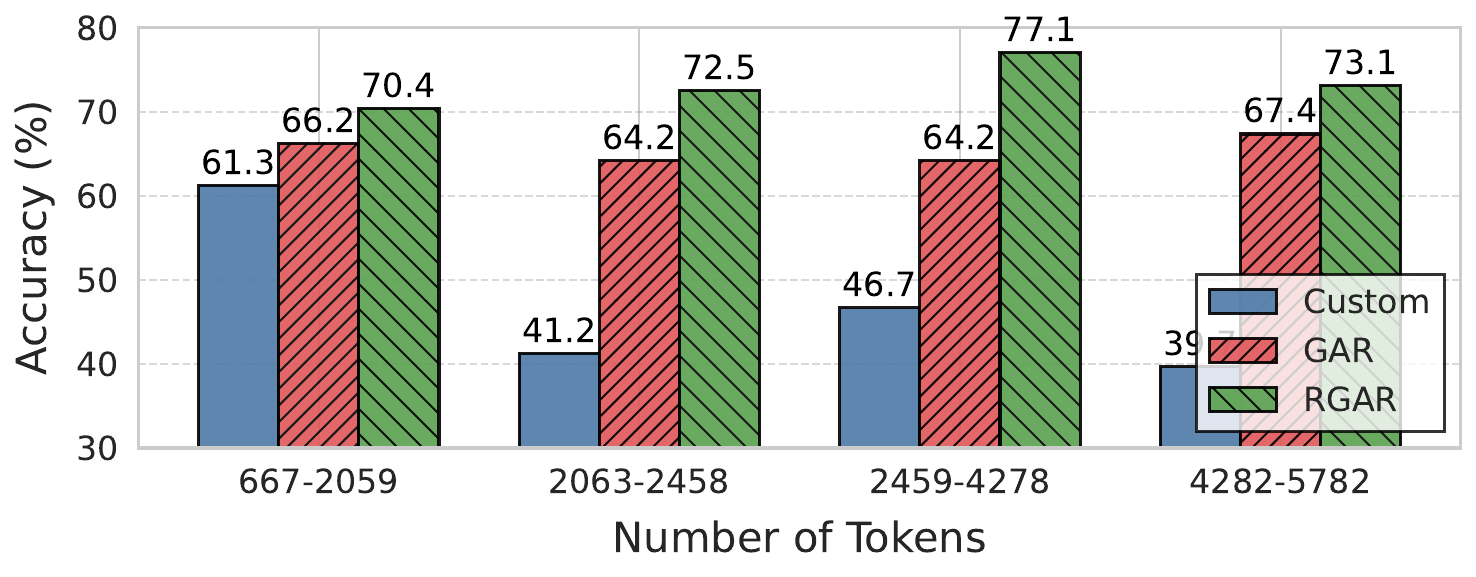}
    \caption{Fine-Grained Accuracy of EHRNoteQA After Sorting by Length and Dividing into Four Equal Parts.}
    \label{fig:bar}
\end{figure}

While the previous sections examined overall dataset performance and established preliminary findings, this section provides a detailed analysis of specific aspects of our results. In § \ref{cross-dataset}, we showed that RGAR performs better on real-world medical recommendation tasks involving comprehensive EHRs. To verify this finding, we conduct a detailed analysis of EHRNoteQA by grouping questions based on context length and dividing them into four bins. Within each bin, we compare the performance of RGAR, GAR, and Custom. As shown in Figure \ref{fig:bar}, Custom shows decreasing accuracy with increasing context length. GAR improves accuracy across all bins, with RGAR achieving further performance gains. Notably, the improvements are more significant in the three bins with longer contexts compared to the first bin. The results show that RGAR maintains consistent average performance across different context length.


\begin{figure}[htbp]
    \centering
    \includegraphics[width=1\linewidth]{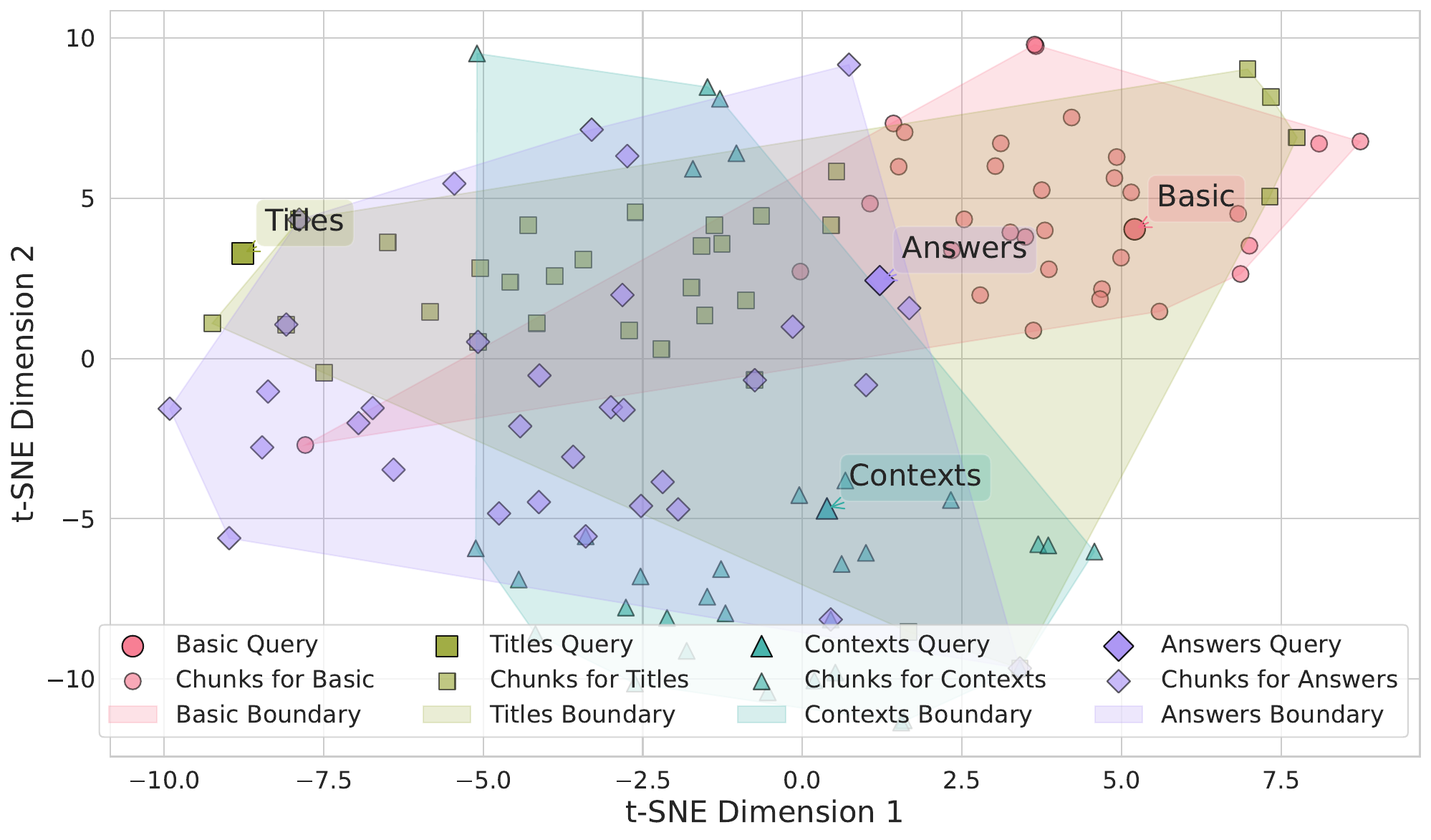}
    \caption{t-SNE Visualization of Different Queries and the Retrieved Chunks.}
    \label{fig:tsne}
\end{figure}

It is also important to note that generating multiple queries from different aspects within RGAR helps stabilize retrieval.
Figure \ref{fig:tsne} presents a t-SNE visualization of different queries and their individually retrieved chunks for a sample question (details provided in Appendix~\ref{case}). The basic query shows limited suitability for retrieval, as its coverage area differs from that of the three queries generated by RGAR. RGAR clearly introduces some variation in retrieval content. Although the regions corresponding to the three generated queries overlap, the specific chunks retrieved do not overlap significantly. This underscores the need to average the retrieval similarities of these three queries to achieve more stable retrieval results.

\section{Conclusion}
In this work, we propose RGAR, a novel RAG system that distinguishes two types of retrievable knowledge. Through comprehensive evaluation across three factual-aware medical benchmarks, RGAR demonstrates substantial improvements over existing methods, emphasizing the significant impact of in-depth factual knowledge extraction and its interaction with conceptual knowledge on enhancing retrieval performance. 
Notably, our RGAR enables the Llama-3.1-8B-Instruct model to outperform the considerably larger, RAG-enhanced proprietary GPT-3.5.
From a broader perspective, RGAR offers a promising approach for enhancing general-purpose LLMs in clinical diagnostic scenarios where extensive factual knowledge is crucial, with potential for extension to other professional domains demanding precise factual awareness. 
\newpage
\section*{Limitations}
Despite RGAR achieving superior average performance, several limitations warrant discussion. Our RGAR requires corpus retrieval, and its time complexity scales proportionally with the size of the corpus, which is an inherent issue within the RAG paradigm. Approaches that generate reasoning evidence directly through domain-specific LLMs \cite{yu2023generate, frisoni-etal-2024-generate} avoid the computational challenges at inference time. However, they face difficulties in updating LLMs to incorporate new medical knowledge, which results in frequent updates and training costs.

Comparative approaches such as MedRAG \cite{xiong-etal-2024-benchmarking} and $i$-MedRAG \cite{xiong2024improving} explore integration possibilities with prompting techniques like Chain-of-Thought \cite{wei2022chain} and Self-Consistency \cite{wang2023selfconsistency} to enhance reasoning capabilities. Our investigation focused specifically on validating how additional factual knowledge processing improves retrieval performance, without examining the impact of these prompting strategies. 
Furthermore, unlike multi-round methods such as $i$-MedRAG \cite{xiong2024improving} that implement LLM-based early stopping to reduce computational costs, our system operates with fixed time complexity. However, it is noteworthy that, because $i$-MedRAG requires multiple rounds of query decomposition, retrieval, and answer aggregation, the actual time overhead of RGAR is significantly smaller than that of $i$-MedRAG.

Our EHR extraction approach assumes LLMs can process complete EHR contextual input, justified by current mainstream LLMs exceeding 128K context windows with anticipated growth. However, in extreme cases where EHR content exceeds LLM context limits, integration with chunk-free approaches may be necessary \cite{luo-etal-2024-landmark, qian-etal-2024-grounding}. Finally, as RGAR operates in a zero-shot setting without instruction fine-tuning, its effectiveness is partially contingent on the model's instruction-following capabilities—which we cannot fully mitigate.

\section*{Ethical Statement}
This research adheres to the ACL Code of Ethics. All medical datasets utilized in this study are either open access or obtained through credentialed access protocols. To ensure patient privacy protection, all datasets have undergone comprehensive anonymization procedures.
While Large Language Models (LLMs) present considerable societal benefits, particularly in healthcare applications, they also introduce potential risks that warrant careful consideration. Although our work advances the relevance of retrieved content for medical queries, we acknowledge that LLM-generated responses based on retrieved information may still be susceptible to errors or perpetuate existing biases.
Given the critical nature of medical information and its potential impact on healthcare decisions, we strongly advocate for a conservative implementation approach. Specifically, we recommend that all system outputs undergo rigorous validation by qualified medical professionals before any practical application. This stringent verification process is essential to maintain the integrity of clinical and scientific discourse and prevent the propagation of inaccurate or potentially harmful information in healthcare settings.
These ethical safeguards reflect our commitment to responsible AI development in the medical domain, where the stakes of misinformation are particularly high and the need for reliability is paramount.

\bibliography{custom}

\appendix

\section{Implementation Details}
\subsection{Hardware Configuration}
All experiments were conducted on an in-house workstation equipped with \textit{dual} NVIDIA GeForce RTX 4090 GPUs, 
128GB RAM, and an Intel® Core i9-13900K CPU.

Time cost across all methods on EHRNoteQA are shown in Table \ref{tab:method_time_comparison}.

\begin{table}[htbp]
  \centering
  \caption{Comparison of different methods in terms of execution time (hours).}
  \resizebox{\linewidth}{!}{
  \begin{tabular}{lccccccc}
    \toprule
    Method & Custom & CoT & RAG & MedRAG & GAR & $i$-MedRAG & RGAR \\
    \midrule
    Time (h) & 0.5 & 0.5 & 1 & 1 & 2 & 22 & 6 \\
    \bottomrule
  \end{tabular}
  }
  \label{tab:method_time_comparison}
\end{table}

\subsection{Code and Results}
The core implementation of the RGAR framework and the output json files can be accessed via the \textbf{Anonymous Repository}: \url{https://anonymous.4open.science/r/RGAR-C613}

\section{Prompt Template and Case Study}
\label{case}
For simplicity, we merged EHR and question in the prompt words of the answer and treated them as question in the prompt words.
Table \ref{tab:prompts} shows the prompts template of RGAR and compared work (Using CoT ones). Table \ref{tab:input} shows the input of a sample, Table \ref{tab:output} shows the final output of RGAR.

\begin{table*}[h]
\centering
\begin{tabular}{p{0.3\textwidth}|p{0.7\textwidth}}
\toprule
Type & Prompt Template \\
\midrule
System prompts for Non-CoT & You are a helpful medical expert, and your task is to answer a multi-choice medical question using the relevant documents. Organize your output in a json formatted as Dict \{"answer\_choice": Str\{A/B/C/...\}\}. Your responses will be used for research purposes only, so please have a definite answer. Please just give me the json of the answer. \\
\midrule
System prompts for using CoT  & You are a helpful medical expert, and your task is to answer a multi-choice medical question. Please first think step-by-step and then choose the answer from the provided options. Organize your output in a json formatted as Dict\{"step\_by\_step\_thinking": Str(explanation), "answer\_choice": Str\{A/B/C/...\}\}. Your responses will be used for research purposes only, so please have a definite answer. Please just give me the json of the answer. \\
\midrule
Answer prompts for Non-CoT &Here are the relevant documents:
\{\{context\}\}
\newline
Here is the question:
\{\{question\}\}
\newline
Here are the potential choices:
\{\{options\}\}
\newline
Please just give me the json of the answer. Generate your output in json:\\

\midrule
Answer prompts for Using CoT &Here are the relevant documents:
\{\{context\}\}
\newline
Here is the question:
\{\{question\}\}
\newline
Here are the potential choices:
\{\{options\}\}
\newline
Please think step-by-step and generate your output in one json:\\
\midrule
Extracting EHR prompts & Here are the relevant knowledge sources:
\{\{context\}\}
\newline
Here are the electronic health records:
\{\{ehr\}\}
\newline
Here is the question:
\{\{question\}\}
\newline
Please analyze and extract the key factual information in the electronic health records relevant to solving this question and present it as a Python list. 
Use concise descriptions for each item, formatted as ["key detail 1", ..., "key detail N"]. Please only give me the list. Here is the list: \\
\midrule
Generating Possible Answer prompts & Please give 4 options for the question. Each option should be a concise description of a key detail, formatted as: A. "key detail 1" B. "key detail 2" C. "key detail 3" D. "key detail 4\\
\midrule
Generating Possible Title prompts & Please generate some titles of references that might address the above question. Please give me only the titles, formatted as: ["title 1", "title 2", ..., "title N"]. Please be careful not to give specific content and analysis, just the title.\\
\midrule
Generating Possible Contexts prompts & Please generate some knowledge that might address the above question. please give me only the knowledge. \\
\bottomrule
\end{tabular}
\caption{Prompt templates used in RGAR and Compared Methods.}
\label{tab:prompts}
\end{table*}

\begin{table*}[h]
\centering
\begin{tabular}{p{0.3\textwidth}|p{0.7\textwidth}}
\toprule
Type & Texts \\
\midrule
EHR & A 39-year-old woman is brought to the emergency department because of fevers, chills, and left lower quadrant pain. Her temperature is 39.1°C (102.3°F), pulse is 126/min, respirations are 28/min, and blood pressure is 80/50 mm Hg. There is blood oozing around the site of a peripheral intravenous line. Pelvic examination shows mucopurulent discharge from the cervical os and left adnexal tenderness. Laboratory studies show: Platelet count 14,200/mm3 Fibrinogen 83 mg/mL (N = 200–430 mg/dL) D-dimer 965 ng/mL (N < 500 ng/mL)\\
\midrule
Questions & When phenol is applied to a sample of the patient's blood at 90°C, a phosphorylated N-acetylglucosamine dimer with 6 fatty acids attached to a polysaccharide side chain is identified. A blood culture is most likely to show which of the following? \\
\midrule
Options & A. Coagulase-positive, gram-positive cocci forming mauve-colored colonies on methicillin-containing agar \newline
B. Encapsulated, gram-negative coccobacilli forming grey-colored colonies on charcoal blood agar \newline
C. Spore-forming, gram-positive bacilli forming yellow colonies on casein agar \newline
\textbf{D. Lactose-fermenting, gram-negative rods forming pink colonies on MacConkey agar} \\
\bottomrule
\end{tabular}
\caption{Case Study of a Sample Input.}
\label{tab:input}
\end{table*}

\begin{table*}[h]
\centering
\begin{tabular}{p{0.3\textwidth}|p{0.7\textwidth}}
\toprule
Type & Texts \\
\midrule
Generated Title & "Identification of Streptococcus pneumoniae in Clinical Specimens",\newline "Phosphorylcholine-Binding Protein A of Staphylococcus aureus",\newline "D-Dimer Assay for Diagnosing Deep Vein Thrombosis",\newline "The Role of Fibrin Degradation Products in Hemostasis",\newline "Infection Control Measures for Hospital-Acquired Bloodstream Infections" \\
\midrule
Generate Context & The description suggests Neisseria gonorrhoeae infection. Neisseria gonorrhoeae produces lipopolysaccharides in its outer membrane, specifically LPS or endotoxin. When heated, these lipopolysaccharides undergo cross-linking reactions resulting in the formation of phosphorylcholine-attached oligomers such as those described in the question stem. These structures can be detected by ELISA using antibodies against specific epitopes on the surface of N. gonorrhoeae cells. The presence of this structure would suggest an active infection due to N. gonorrhoeae. Blood cultures for N. gonorrhoeae typically require specialized media containing antibiotics like vancomycin, colistin, and nystatin to inhibit competing bacteria \\
\midrule
Generate Answer & Option 1: Streptococcus pneumoniae The presence of a high fever, tachycardia, hypotension, and low platelets suggests sepsis or severe infection. The laboratory findings also indicate disseminated intravascular coagulation (DIC). The identification of a phosphorylated N-acetylglucosane dimer with 6 fatty acid chains attached to a polysaccharide side chain after applying phenol at 90°C is consistent with pneumococcal capsular polysacarcharides.\newline
Option 2: Staphylococcus aureus Staphylococci can cause various infections but do not typically produce this type of capsule structure when exposed to heat.\newline
Option 3: Escherichia coli E. coli does have capsules that may be detected by certain methods such as latex agglutination tests; however, they don't match the description given in the prompt.\newline
Option 4: Neisseria meningitidis Neisseria species like N. meningitidis would more commonly present with symptoms related to central nervous system involvement rather than pelvic signs described in the scenario.\\
\midrule
Extract EHR & ['Patient age: 39 years', 'Symptoms: fever, chills, left lower quadrant pain', 'Vital signs: \textbf{high} temperature (39.1°C), tachycardia (126/min), tachypnea (28/min), hypotension (80/50 mmHg)', 'Physical exam findings: mucopurulent discharge from cervix, left adnexal tenderness', 'Laboratory results: \textbf{low} platelet count (14,200/mm\^{}3), \textbf{elevated} D-dimer (965 ng/mL)', 'Phenol test result: phosphorylated N-acetylglucosame dimmer with 6 fatty acids attached to a polysaccharide side chain'] \\
\midrule
Output Json & \{"answer\_choice": "D"\}\\
\bottomrule
\end{tabular}
\caption{Case Study of output in RGAR.}
\label{tab:output}
\end{table*}

\section{Framework Insight}
\subsection{Another View of the Recurrence Pipeline}
We conceptualize the Recurrence Pipeline as an exploration-exploitation process within the reinforcement learning framework \cite{10.1023/A:1013689704352}. In GAR, even when generated content is only partially accurate (or potentially inaccurate), it remains valuable for retrieval if it correlates with passages containing correct information (e.g., co-occurrence with correct answers), thus representing an exploratory phase. Conversely, EHR extraction serves as an exploitation phase, thoroughly utilizing explored knowledge by selecting relevant components and synthesizing new evidence (factual knowledge). Based on this newly derived evidence, subsequent iterations can initiate fresh exploration-exploitation cycles, creating a continuous knowledge transmission process \cite{10446501}.

In scenarios where additional factual knowledge is not required, the retrieved content tends to remain relatively constant, and utilizing this content under identical prompting conditions would likely yield similar factual knowledge through extraction and summarization. However, when conceptual knowledge is needed to derive new factual knowledge through reasoning from existing factual information, the updated basic query facilitates easier retrieval of conceptual knowledge supporting current reasoned factual knowledge, thereby maintaining the integrity of reasoning chains. Furthermore, leveraging current factual knowledge for retrieval enables the exploration and discovery of novel knowledge domains.

\subsection{Why No Flexible Stopping Criteria}
Similar multiround RAG systems have adopted more flexible stopping criteria. For instance, Adaptive RAG \cite{jeong-etal-2024-adaptive} determines whether to retrieve further 
by consulting the model itself. $i$-MedRAG \cite{xiong2024improving}, while setting a maximum number of retrieval iterations, also supports early stopping.

In our RGAR framework, we do not adopt such settings. On the one hand, we focus on evaluating how additional processing of \textit{factual knowledge} enhances retrieval performance, raising awareness of this often-overlooked type of knowledge in previous RAG systems, while flexible stopping criteria mainly showcase procedural knowledge and metacognitive knowledge. On the other hand, the metacognitive capabilities of current LLMs remain under question, as a model’s self-evaluation of the need for additional retrieval information often does not match actual requirements \cite{kumar-etal-2024-confidence}.

\subsection{Future Work}
Our RGAR framework leverages retrieved medical domain knowledge to deliver exceptional answer quality
. However, we are concerned that such powerful generative capabilities, if maliciously exploited, could pose security risks. For instance, when the retrieved corpus contains private or copyrighted information, malicious users could exploit the LLM's responses to extract and disclose sensitive data from the corpus \cite{carlini2021extracting}. 
Additionally, malicious users might attempt to replicate our base LLM \cite{tramer2016stealing, zhu2024efficient} by collecting large volumes of question-answer pairs or infer internal details of our retrieval-based generation framework \cite{carlinistealing}. 
We will make every effort to mitigate these risks, such as verifying the legitimacy of queries \cite{inan2023llama} and watermarking the models used in RGAR \cite{zhu2024reliable}.
, ensuring that RGAR is used responsibly and legally.

\section{Comparative Analysis of Dataset Length Distributions}
\begin{figure*}[htbp]
        \centering
    \includegraphics[width=1\linewidth]{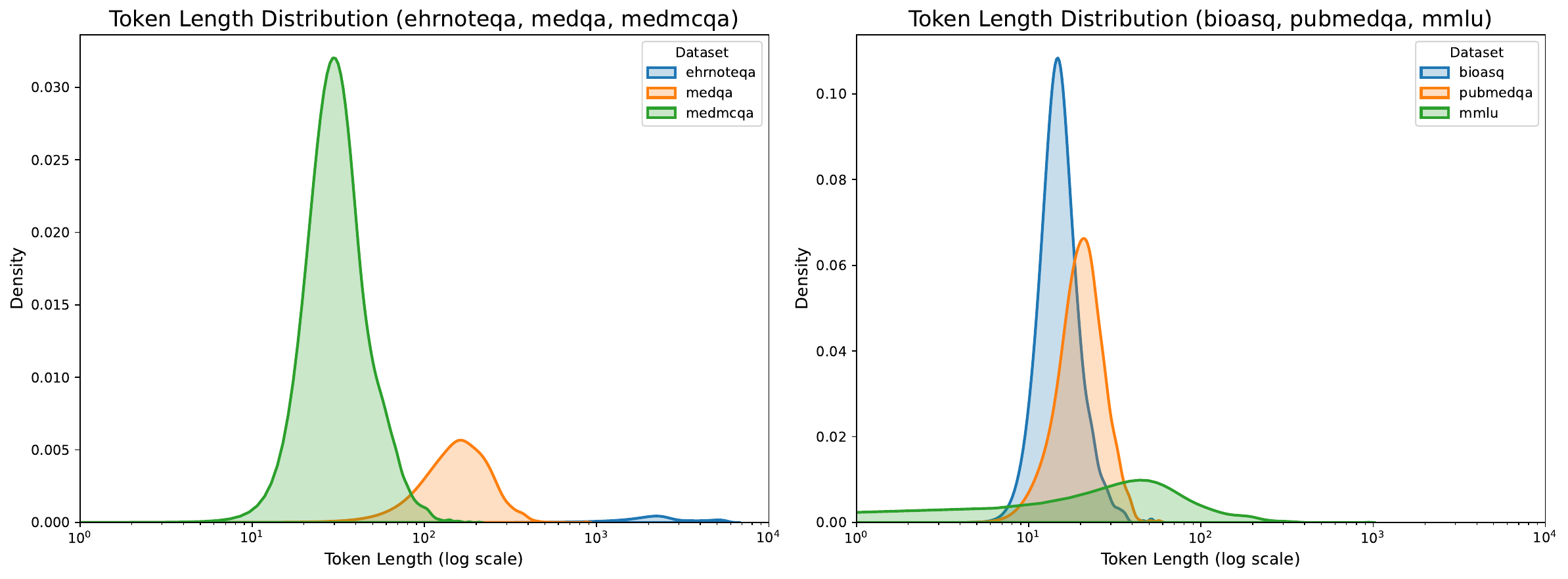}
    \caption{Length Distribution Analysis of Medical QA Datasets with and without EHR.}
    \label{fig:kde}
\end{figure*}

In this section, we present additional visualizations comparing the two categories of datasets we described, and explain our rationale for excluding the MMLU-med dataset \cite{hendrycks2021measuring}. We plotted smoothed Kernel Density Estimation (KDE) curves for these datasets, as shown in Figure \ref{fig:kde}. Our analysis confirms that datasets containing Electronic Health Records (EHR) consistently demonstrate greater length compared to those without EHR content. However, certain datasets exhibit complex question sources and types. For instance, while the MMLU dataset shows a considerable mean length of 84 tokens and a maximum length of 961 tokens, as illustrated in the figure, the vast majority of its questions lack EHR content and are predominantly shorter in length. This characteristic led to our decision to exclude it from our experimental evaluation.
\ref{fig:kde}
\end{document}